\documentclass{article}

\usepackage{hyphenat}

\usepackage{PRIMEarxiv}

\usepackage[utf8]{inputenc} 
\usepackage[T1]{fontenc}    
\usepackage{hyperref}       
\usepackage{url}            
\usepackage{booktabs}       
\usepackage{amsfonts}       
\usepackage{nicefrac}       
\usepackage{microtype}      
\usepackage{lipsum}
\usepackage{fancyhdr}       
\usepackage{graphicx}       
\graphicspath{{media/}}     

\title{PACE: Publisher-Adaptive Content Extraction via Agentic Automation
\thanks{\textit{\underline{Citation}}: 
\textbf{Authors. Title. Pages.... DOI:000000/11111.}} 
}

\author{
  Zhanlin Liu, Munirathnam Srikanth \\
  ProRata.ai \\
  Seattle\\
  \texttt{\{zhanlin, srikanth\}@prorata.ai} \\
}

\begin{document}
\maketitle

\begin{abstract}
Web content extraction is essential for reliable LLM data pipelines, yet
existing methods often struggle to jointly satisfy accuracy, scalability,
and adaptability. General-purpose extractors can be applied broadly, but
they are often brittle on publisher-specific layouts and richer
extraction targets such as metadata, images, and tables. Direct
LLM-based extraction offers greater flexibility, but incurs substantial
cost and latency at scale, while manually engineered
publisher-specific parsers can achieve high accuracy but require
substantial human effort to build and maintain.

We introduce PACE, an agentic framework for learning
publisher-specific extraction configurations from representative pages
and user requirements. During training, PACE uses LLMs to analyze page
structure and aggregate reusable extraction patterns. At inference time,
the learned configurations instantiate a fixed deterministic extractor
template, enabling scalable extraction without additional LLM calls.

Experiments spanning article-body, metadata, and multimodal extraction
show that PACE outperforms scalable non-manual baselines while
approaching the quality of manually engineered publisher-specific
parsers. PACE achieves stronger extraction of article text, metadata,
images, and tables, demonstrating that agentic configuration learning can
automate publisher-specific extraction for LLM-ready page
representations beyond article text.
\end{abstract}


\section{Introduction}

Web content extraction is a foundational step in building high-quality
data pipelines for search, retrieval-augmented generation (RAG),
summarization, question answering, and other large language model (LLM)
applications~\cite{kohlschutter2010boilerplate,pomikalek2011removing,
lewis2020retrieval,gao2023retrieval,li2026beyond}. Although web pages
contain rich information, they are primarily designed for human
presentation rather than machine-readable processing. Consequently,
target content is often interleaved with navigation menus,
advertisements, recommendation modules, comments, embedded media,
structured tables, captions, metadata, and publisher-specific layout
templates~\cite{kohlschutter2010boilerplate,pomikalek2011removing}.
Effective extraction therefore requires separating semantically relevant
information from boilerplate, presentation-specific structures, and
other non-content artifacts~\cite{barbaresi2021trafilatura}.

The growth of LLM applications increases both the importance and the
difficulty of web content extraction. Modern LLM pipelines increasingly
require more than clean article text: they often depend on task-specific
and multimodal page representations, including metadata, tables,
figures, images, captions, links, and other structured
elements~\cite{li2026beyond,liu2023visual,zhao2023retrieving,
mei2025survey}. At the same time, extraction requirements vary across
downstream tasks. A retrieval system may require plain article text, a
multimodal grounding pipeline may require markdown with images and
captions, and an information extraction or tool-use system may require
structured JSON containing metadata, tables, and links. Thus, web content
extraction is not merely a boilerplate-removal problem, but a
requirement-adaptive data-understanding problem.

Existing extraction systems face a persistent tradeoff among accuracy,
scalability, and adaptability. General-purpose heuristic and
machine-learning-based methods can be deployed across many websites, but
they typically rely on fixed assumptions about page structure, text
density, link density, readability signals, or training
distributions~\cite{kohlschutter2010boilerplate,pomikalek2011removing,
barbaresi2021trafilatura,hamborg2017news,boilernet}. Consequently, they
may struggle with publisher-specific templates, interactive pages, and
extraction targets outside their predefined schemas. Direct LLM-based
extraction offers greater flexibility because extraction goals can be
specified in natural language, but invoking an LLM for every page
introduces substantial costs, latency, and potential
nondeterminism~\cite{xu2024large}. Manually engineered
publisher-specific systems such as Fundus demonstrate that customized
parsers can achieve high extraction quality, but they require manual DOM
inspection, rule authoring, and ongoing
maintenance~\cite{dallabetta2024fundus}.

The effectiveness of publisher-specific systems points to an important
opportunity: once a publisher's recurring DOM structure and layout
conventions are understood, a customized extractor can be more reliable
than a one-size-fits-all approach~\cite{dallabetta2024fundus}. However,
this customization is typically performed manually, requiring substantial
engineering effort. Recent advances in LLM reasoning and agentic
workflows provide a path toward automating this process. LLMs can
interpret natural-language requirements and reason over semi-structured
inputs, while agentic workflows can coordinate analysis, intermediate
artifact generation, feedback, and refinement~\cite{yao2023react,yao2023tree,
besta2024graph,zhao2024expel,wu2024autogen}.

This observation motivates a different design principle: use LLMs to
learn publisher-specific extraction behavior while keeping the deployed
extractor deterministic and reusable. We propose PACE:
Publisher-Adaptive Content Extraction via Agentic Automation, a framework
that learns extraction configurations from representative pages and
user-defined requirements. PACE uses an agentic orchestration layer to
construct prompts, guide page-level DOM analysis, aggregate recurring
publisher-level patterns, incorporate optional human feedback, and
produce reusable publisher-specific configurations.

Rather than generating unconstrained extractor code, PACE uses each
learned configuration to instantiate a fixed deterministic extractor
template. The template handles DOM traversal, inclusion and exclusion
logic, fallback extraction, deduplication, normalization, and output
formatting, while the configuration specifies publisher-specific
selectors and rules. This design preserves the adaptability of
LLM-guided analysis during training while enabling scalable inference
without additional LLM calls.

In summary, PACE makes three contributions. First, it formulates
publisher-specific web extraction as a configuration-learning problem,
replacing manual parser construction with agentic analysis of
representative pages and user requirements. Second, it separates
LLM-guided training from template-based inference: LLMs learn reusable
extraction configurations, while a fixed extractor template performs
scalable extraction without per-page LLM calls. Third, it provides
empirical evidence across article text, metadata, and multimodal
extraction, showing that PACE achieves the strongest performance among
evaluated scalable non-manual extractors while approaching the quality of
manually engineered publisher-specific parsers. These results demonstrate
that publisher-adaptive configuration learning can support flexible
LLM-ready page representations beyond article text.

\section{Related Work}

Web content extraction has been studied from several perspectives,
including heuristic boilerplate removal, machine-learning-based
extraction, HTML-to-text conversion for LLM data pipelines,
publisher-specific parsing, and recent LLM-based extraction workflows.
These approaches differ in how they balance accuracy, scalability, and
adaptability. PACE connects these lines of work by using LLMs and
agentic workflows to automate publisher-specific configuration learning,
while retaining deterministic inference through a fixed extractor
template.

\subsection{Heuristic and Rule-Based Extraction}

Early web content extraction systems separate the main content from
boilerplate using structural and textual signals. Boilerpipe uses shallow
text features to identify content-bearing text blocks
~\cite{kohlschutter2010boilerplate}, while jusText removes boilerplate
from web corpora using text-density and link-density features
~\cite{pomikalek2011removing}. More recent systems, such as trafilatura
~\cite{barbaresi2021trafilatura} and news-please
~\cite{hamborg2017news}, combine multiple heuristics and extraction
strategies to support general-purpose web and news extraction.

These systems are broadly applicable and computationally efficient, but
their extraction behavior is largely fixed. They typically assume that
target content resembles conventional article text and that boilerplate
can be detected from density, link structure, or readability signals. As
a result, they may struggle with publisher-specific templates,
interactive pages, and user-defined extraction targets such as
structured metadata, in-content images, captions, and tables. In
contrast, PACE learns publisher-specific extraction configurations from
representative pages and user requirements, allowing the same framework
to adapt to both article-body extraction and richer LLM-ready page
representations.

\subsection{Machine-Learning-Based Boilerplate Removal}

Machine-learning-based methods reduce reliance on manually designed
heuristics by learning to distinguish content regions from boilerplate.
BoilerNet formulates boilerplate removal as neural sequence labeling over
HTML text blocks~\cite{boilernet}. Such methods can learn richer
structural and textual patterns than fixed heuristics, but their behavior
remains constrained by the training distribution and the task definition.

This limitation creates challenges when deploying learned extractors
across new publishers and evolving web templates. A model trained for
boilerplate removal may not transfer reliably to layouts or extraction
targets that differ from those seen during training, particularly when
the target expands beyond article-body text to metadata, structured
tables, captions, or multimodal page elements. PACE addresses this issue
by learning publisher-specific configurations at deployment time from a
small set of representative pages, rather than relying solely on a
globally trained extractor.

\subsection{HTML-to-Text Extraction for LLM Data}

Recent work has revisited HTML-to-text extraction in the context of LLM
data pipelines. Li et al. show that relying on a single extractor is
poorly suited to the diversity of web content used for LLM training, and
that extractor choice affects both data coverage and downstream model
behavior~\cite{li2026beyond}. This perspective highlights that
extraction is not a neutral preprocessing step: the extractor determines
which information enters the LLM data pipeline and which information is
discarded.

PACE is complementary to this line of work. Rather than selecting a
single fixed extractor or combining the outputs of existing extractors,
PACE learns publisher-specific configurations conditioned on user
requirements. This shifts the focus from choosing an extractor to
automatically adapting extraction behavior to a publisher, page template,
and target schema. This adaptation is especially important when LLM
pipelines require representations beyond plain text, such as markdown
with images, structured metadata, or preserved table content.

\subsection{LLM-Based Information Extraction}

LLMs provide a flexible interface for information extraction because
schemas and extraction requirements can be specified in natural language.
Surveys of generative information extraction highlight the ability of
LLMs to perform schema-guided and instruction-following extraction
across a wide range of tasks~\cite{xu2024large}. This flexibility is
valuable for web extraction, where applications may require different
fields, output formats, or multimodal page elements.

However, using an LLM as the extractor for every page introduces
recurring inference costs, latency, and potential nondeterminism. For many
publishers, page layouts are repetitive: once the DOM structure is
understood, the same extraction behavior can be reused across many pages.
PACE therefore uses LLMs during training to infer reusable
publisher-specific configurations, and then performs inference with a
deterministic extractor template. This design preserves the adaptability
of LLM-guided extraction while avoiding per-page LLM calls at scale.

\subsection{LLM Agents and Workflow Automation}

PACE is also related to recent work on LLM agents and automated
workflows. ReAct shows that LLMs can interleave reasoning and actions when
interacting with external tools or environments~\cite{yao2023react}.
The Tree of Thoughts and Graph of Thoughts study structured reasoning over
intermediate thought states~\cite{yao2023tree,besta2024graph}. ExpeL
shows that LLM agents can learn from accumulated experience without
parameter updates~\cite{zhao2024expel}. AutoGen demonstrates that
multi-agent LLM systems can coordinate conversation, tool use, and human
input to solve complex tasks~\cite{wu2024autogen}. These works
motivate the use of LLMs not only as direct predictors but also as
workflow participants that can analyze examples, produce intermediate
artifacts, incorporate feedback, and refine outputs.

PACE applies this agentic perspective to web content extraction. The
framework uses an agentic orchestrator to interpret requirements,
construct prompts, coordinate page-level analysis and publisher-level
aggregation, incorporate optional feedback, and refine extraction
configurations. Unlike general-purpose agent frameworks, PACE targets a
specific production-oriented problem: converting LLM-guided structural
analysis into reusable deterministic extractors.

\subsection{Publisher-Specific Extraction}

Publisher-specific extraction systems show that customized parsers can
achieve high-quality extraction when the target publisher is known.
Fundus, for example, uses manually engineered parsers for individual
publishers and is optimized for high-quality news extraction
~\cite{dallabetta2024fundus}. This suggests that publisher-specific
extraction is an effective abstraction: once a publisher's recurring
layout patterns are captured, extraction can be more reliable than with a
general-purpose extractor.

The primary limitation is scalability. Manually inspecting DOM
structures, writing extraction rules, and maintaining publisher-specific
parsers requires substantial engineering effort, especially as publishers
update templates or as extraction requirements expand beyond article
text. PACE addresses this limitation by automating the customization
process. It retains the benefits of publisher-specific extraction while
replacing manual parser development with agentic configuration learning
and template-based deterministic inference.

The next section formalizes this configuration-learning approach and
describes how PACE converts representative pages and user requirements
into deterministic publisher-specific extractors.

\section{Methodology}

PACE separates publisher-specific configuration learning from extraction
execution. Given a publisher $p$, a small set of representative pages
$\mathcal{D}_{p}^{train}$, and user requirements $r$, PACE learns a
publisher-level extraction configuration $c_p$. This configuration
specifies reusable selectors, exclusion rules, fallback logic, and
output-format instructions. PACE then uses $c_p$ to instantiate a fixed
deterministic extractor template, which is applied to new pages from the
same publisher without additional LLM calls.

\begin{figure*}[t]
    \centering
    \includegraphics[width=0.95\linewidth]{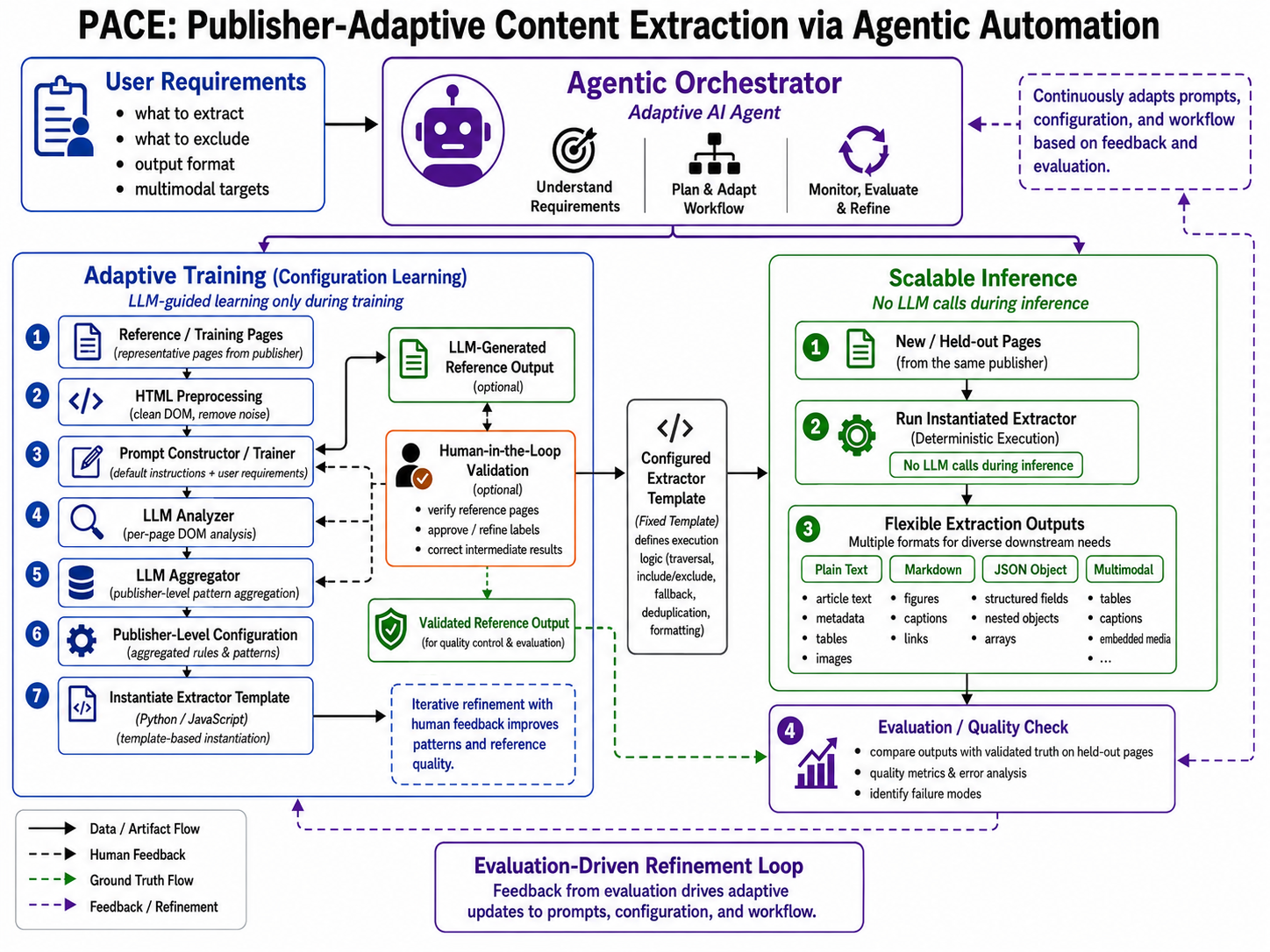}
\caption{
PACE architecture. The agentic orchestrator first interprets user
requirements and coordinates training-time analysis over representative
publisher pages. During training, cleaned HTML pages are analyzed by an
LLM-based page-level analyzer to produce extraction specifications,
which are then aggregated into a publisher-level configuration. Optional
human feedback and evaluation results can be used to refine prompts,
rules, and configurations. At inference time, the learned configuration
instantiates a fixed extractor template, which applies publisher-specific
selectors, exclusion rules, fallback logic, and formatting instructions
to new pages without additional LLM calls. The resulting extractor can
produce text, structured metadata, tables, images, captions, markdown,
JSON, or other task-specific outputs.
}
    \label{fig:pace_architecture}
\end{figure*}

Figure~\ref{fig:pace_architecture} illustrates the overall architecture.
An agentic orchestrator interprets user requirements, coordinates
training-time LLM analysis, incorporates optional human feedback, and
uses evaluation results to refine prompts, configurations, and workflow
decisions. The central design choice is to use LLMs for configuration
learning rather than per-page extraction. This preserves the adaptability
of LLM-guided analysis during training while enabling scalable
template-based inference without additional LLM calls. In our implementation, the page-level analyzer and publisher-level
aggregator are instantiated with an LLM; the specific model used in the
experiments is reported in Section Experiments.

\subsection{Problem Formulation}

For each publisher $p$, PACE is given a set of representative training
pages $\mathcal{D}_{p}^{train} = \{x_1, x_2, \ldots, x_n\}$ and a set of
user requirements $r$. Each page $x_i$ consists of raw HTML containing
both target content and non-content regions, such as navigation menus,
advertisements, recommendation modules, comments, and layout-specific
boilerplate. The requirements $r$ define the extraction target, excluded
content, desired output format, and optional structured or multimodal
fields.

The objective is to learn a reusable publisher-level configuration
$c_p$ that generalizes from the representative pages to unseen pages from
the same publisher. Given a new page $x$, PACE applies a
publisher-specific extractor $f_p$, instantiated from $c_p$, to produce
the extracted output:

\[
f_p(x; c_p, r) \rightarrow y,
\]

where $y$ denotes the extracted representation. Depending on the
requirements $r$, $y$ may be plain text, markdown, a JSON object,
metadata fields, tables, images, captions, or another downstream
representation.

\subsection{Agentic Configuration Learning}

PACE learns $c_p$ through an adaptive training pipeline coordinated by
the agentic orchestrator. Given user requirements, the orchestrator
constructs prompts for two training-time LLM components: a page-level
analyzer and a publisher-level aggregator. The analyzer infers
extraction-relevant structure from individual pages, while the aggregator
consolidates page-level observations into a reusable publisher-level
configuration.

Before LLM analysis, PACE preprocesses each reference page to reduce DOM
complexity. Scripts, styles, hidden elements, and common structural noise
are removed, while visible content structure and text-bearing elements
are preserved. This produces a simplified DOM representation that allows
the analyzer to focus on extraction-relevant page structure.

For each preprocessed page, the page-level analyzer produces a structured
extraction specification. This specification may include candidate
selectors for the content root, body text, title, metadata fields,
paragraphs, structured elements, and regions to exclude. The analyzer is
prompted to prefer selectors that are likely to generalize across pages
from the same publisher and to avoid overly specific rules tied to a
single article. Unlike direct LLM extraction, this step does not return
the final extracted content; instead, it produces reusable structural
knowledge for extracting the requested fields from pages with similar
layouts.

The publisher-level aggregator then compares page-level specifications
across the reference pages. It identifies stable DOM patterns, removes
one-off or overly specific selectors, generalizes unstable class names,
and resolves conflicts among candidate rules. For example, CSS-module
suffixes or hash-like class patterns can be converted into more robust
partial-match selectors. The output is a publisher-level configuration
$c_p$ that specifies how to locate target content, remove noise, apply
fallback logic, and format the extracted output. This configuration is
structured to match the fields expected by the extractor template,
enabling controlled template instantiation rather than free-form code
generation.

\subsection{Template-Based Extractor Generation and Inference}

After configuration learning, PACE instantiates an extractor by applying
the publisher-level configuration $c_p$ to a reusable code template. This
is not unconstrained code generation: the executable extraction logic is
defined by a fixed template, while the learned configuration supplies
publisher-specific selectors, exclusion rules, fallback rules, and
field-formatting instructions.

The template defines the general extraction procedure, including DOM
traversal, selector application, inclusion and exclusion logic, fallback
extraction, deduplication, normalization, and output formatting. The
configuration determines which DOM regions and fields are selected for a
given publisher and requirement set. This separation improves
controllability and reproducibility: changes to publisher-specific
extraction behavior are localized in $c_p$, while the runtime logic
remains fixed and auditable.

During inference, the instantiated extractor is applied to new or
held-out pages from the same publisher. The page is loaded, the learned
selectors and exclusion rules are executed, boilerplate is removed, and
the requested content is returned in the specified format. Because the
extractor is a template-based instantiation rather than a per-page LLM
prediction, inference does not require additional LLM calls and avoids
the cost, latency, and nondeterminism of direct LLM extraction.

The same inference pipeline can support different output targets because
$c_p$ is conditioned on user requirements. In our experiments, we
evaluate both article-body extraction and broader metadata and
multimodal extraction, including metadata fields, images, and tables.
The same framework can also represent requirements for captions,
embedded media, markdown, JSON, or other structured outputs.

\subsection{Human Feedback and Evaluation-Driven Refinement}

PACE supports optional human feedback during configuration learning. In
settings where manually curated gold labels are unavailable, the system
can generate candidate reference outputs from representative pages in a
human-readable format such as markdown. Users may review these outputs,
verify whether the reference pages are suitable, correct labels, refine
requirements, or inspect intermediate analyzer and aggregator outputs.
This feedback can be incorporated into prompt construction, page-level
analysis, and publisher-level aggregation.

PACE also includes an evaluation-driven refinement loop. When validated
references are available, extraction outputs from held-out pages are
compared against reference outputs using task-specific metrics. In the
article-body extraction benchmark, extracted text is compared against
manually verified paragraph arrays using ROUGE-LSum and word error rate.
Evaluation feedback can reveal failure modes such as missing article
segments, excessive boilerplate, layout-specific errors, missing
captions, or overly aggressive noise removal. The orchestrator can then
revise prompts, update aggregation rules, request additional
representative pages, or trigger further human validation.

Candidate reference generation is optional in the training workflow and
is used for inspection or human validation when available. In the
metadata and multimodal benchmark, GPT-5.4-generated references are used
only as evaluation references for held-out pages and are not used to
train PACE. The prompts used for page-level rule extraction and
publisher-level aggregation, along with the HTML preprocessing and
extractor-template implementation, are provided in the supplementary
material.

\section{Experiments}
\label{sec:experiments}

We evaluate PACE in two complementary settings. The first evaluates
article-body extraction using manually verified annotations from the
Fundus benchmark. This setting provides controlled evidence for
main-text extraction quality and cross-publisher robustness. The second
evaluates metadata and multimodal extraction, covering metadata fields,
in-content images, and tables. Because manually labeled ground truth is
not available for these broader targets, we construct an LLM-generated
reference benchmark: GPT-5.4 extracts canonical metadata and markdown
content from each evaluation page, and both PACE and trafilatura are
scored against the same reference. For image extraction, we use a
deterministic reference derived from in-content image URLs in the raw
HTML, since GPT-5.4 under-extracts images.

In all experiments, GPT-5.4 is used for PACE's training-time LLM
components, including page-level analysis and publisher-level
aggregation. The instantiated extractors are then applied to held-out
pages without additional LLM calls. Together, these settings test whether
PACE can achieve high-quality article-body extraction while also
supporting richer LLM-ready page representations through reusable
publisher-specific extractors.

\subsection{Datasets}

The article-body benchmark is constructed from the Fundus news extraction
dataset, which provides manually verified body-paragraph annotations.
This benchmark contains 16 news publishers and 71 held-out evaluation
pages, covering diverse publisher templates, article layouts, and
boilerplate patterns.

PACE requires a small set of representative pages from each publisher to
learn publisher-level extraction configurations. For 13 publishers, we
crawl five additional pages per publisher using Playwright. These
crawled pages are separate from the Fundus benchmark and are used only
for configuration learning; all available Fundus pages for these
publishers are reserved for held-out evaluation. For the remaining three
publishers, additional crawling failed because of CAPTCHA or access
restrictions. We therefore use a fallback split of the available Fundus
pages, using three pages for configuration learning and two pages for
held-out evaluation. In all cases, training pages are used only to infer
publisher-specific configurations, while held-out pages are used only for
final measurement.

The metadata and multimodal benchmark extends the evaluation beyond the main
article text. It contains two page groups. The first reuses the 16 news
publishers from the article-body benchmark to evaluate metadata and
in-content image extraction. The second adds five table-rich domains with
consistent page templates: Wikipedia list pages, StockAnalysis financial
pages, Basketball Reference standings pages, BLS employment-release
pages, and W3Schools reference pages. Each evaluation page is represented
using a shared extraction schema covering metadata fields, markdown
content, in-content images, and tables. For each table-rich domain, we
crawl five training pages and five held-out evaluation pages. In total,
this benchmark covers 21 publishers or domains, approximately 99
training pages, and approximately 96 held-out evaluation pages.

\subsection{Baselines}

For article-body extraction, we compare PACE against seven open-source
extraction systems. Fundus serves as a manually engineered
publisher-specific reference baseline, using hand-crafted parsers for
individual publishers. The remaining systems represent scalable
non-manual approaches: trafilatura combines heuristic and
machine-learning-based extraction; news-please is a general-purpose news
extraction pipeline; BTE and Boilerpipe use shallow text features;
BoilerNet uses neural boilerplate removal; and jusText relies on
text-density and link-density heuristics.

For metadata and multimodal extraction, we compare PACE with
trafilatura. We choose trafilatura because it is the strongest scalable
non-manual baseline in the article-body benchmark and provides support
for metadata, images, and tables through its markdown extraction
interface. PACE is trained separately for each publisher or domain using
the corresponding training pages and is evaluated on held-out pages.
trafilatura is used as a fixed, untrained extractor. 

\subsection{Evaluation Metrics}

For article body extraction, we concatenate the extracted body paragraphs and
compare them with the concatenated ground-truth article body. We report
ROUGE-LSum precision, recall, and F1, as well as word error rate (WER).
ROUGE-LSum~\cite{lin2004rouge} measures longest common subsequence
overlap between extracted and reference text. Precision measures the
fraction of extracted content that corresponds to the reference, recall
measures the fraction of reference content that is recovered, and F1
balances the two. WER measures normalized word-level edit distance and is commonly used to
compare a hypothesis text sequence against a reference sequence
~\cite{levenshtein1966binary,morris2004wer}. In our setting, insertions
generally indicate boilerplate or non-article text, while deletions
indicate missing article content. To
assess robustness across publishers, we also report the standard
deviation of ROUGE-L F1, average per-publisher rank, and the number of
publishers on which each method ranks in the top two.

For metadata and multimodal extraction, both PACE and trafilatura emit a
common JSON schema containing metadata fields and markdown content. The
metadata fields include title, author, publication date, modified date,
URL, description, language, canonical URL, site name, section, and tags.
Metadata accuracy is computed over the reference fields that are present
for each page. We use exact or normalized matching for fields such as
title, language, canonical URL, site name, section, and dates; fuzzy
matching for author and description; and set overlap for tags. Image
extraction is evaluated using URL-set precision, recall, and F1 against
the deterministic in-content image reference. Table extraction is
evaluated using cell-set precision, recall, and F1, where each table is
represented as a set of column-index and cell-text pairs. This makes the
metric robust to formatting differences between markdown and HTML while
still measuring whether table content is preserved.

\subsection{Results}

Table~\ref{tab:multimodal_results} reports metadata, image, and table
extraction performance across the 21-publisher metadata and multimodal
benchmark. After evaluating main-text extraction, this benchmark assesses
whether PACE can preserve structured and multimodal page elements needed
for LLM data pipelines. PACE outperforms trafilatura on all three
targets, achieving 0.755 metadata accuracy versus 0.574, 0.516 image F1
versus 0.318, and 0.929 table F1 versus 0.461. The largest difference
appears in table extraction, where PACE more than doubles trafilatura's
F1.

\begin{table*}[t]
\centering
\small
\setlength{\tabcolsep}{4pt}
\begin{tabular}{lcccccc}
\hline
\textbf{Method} & 
\textbf{ROUGE-L F1} $\uparrow$ & 
\textbf{Precision} $\uparrow$ & 
\textbf{Recall} $\uparrow$ & 
\textbf{WER} $\downarrow$ &
\textbf{Avg. Rank} $\downarrow$ &
\textbf{Top-2} $\uparrow$ \\
\hline
Fundus & \textbf{0.9804} $\pm$ 0.0458 & \textbf{0.9993} $\pm$ 0.0034 & 0.9653 $\pm$ 0.0717 & \textbf{0.0374} $\pm$ 0.0709 & \textbf{1.75} & \textbf{12/16} \\
PACE & 0.9758 $\pm$ \textbf{0.0447} & 0.9773 $\pm$ 0.0409 & 0.9782 $\pm$ 0.0666 & 0.0521 $\pm$ 0.0781 & 2.44 & 11/16 \\
trafilatura & 0.9541 $\pm$ 0.1046 & 0.9298 $\pm$ 0.1336 & 0.9821 $\pm$ 0.0361 & 0.1689 $\pm$ 0.6937 & 2.88 & 5/16 \\
news-please & 0.9347 $\pm$ 0.1121 & 0.9801 $\pm$ 0.1052 & 0.9117 $\pm$ 0.1179 & 0.1955 $\pm$ 0.7968 & 4.00 & 4/16 \\
jusText & 0.8838 $\pm$ 0.1711 & 0.8893 $\pm$ 0.1560 & 0.9081 $\pm$ 0.1779 & 0.2583 $\pm$ 0.4273 & 6.12 & 0/16 \\
BTE & 0.8749 $\pm$ 0.1537 & 0.8112 $\pm$ 0.2014 & \textbf{0.9896} $\pm$ 0.0242 & 0.4203 $\pm$ 0.8738 & 5.69 & 1/16 \\
BoilerNet & 0.8733 $\pm$ 0.1539 & 0.8687 $\pm$ 0.1636 & 0.9203 $\pm$ 0.1599 & 0.3358 $\pm$ 0.6680 & 6.31 & 0/16 \\
Boilerpipe & 0.8029 $\pm$ 0.2654 & 0.8325 $\pm$ 0.2171 & 0.8238 $\pm$ 0.3006 & 0.3453 $\pm$ 0.4046 & 6.25 & 0/16 \\
\hline
\end{tabular}
\caption{
Article-body extraction performance and cross-publisher robustness across
71 held-out pages. Avg. Rank is the average per-publisher rank based on
ROUGE-L F1, with ties assigned the best shared rank. Top-2 counts the
number of publishers on which each method ranks first or second. Fundus
is a manually engineered publisher-specific reference baseline. PACE
achieves the strongest ROUGE-L F1 and WER among scalable non-manual
methods, while remaining close to Fundus and ranking in the top two on
11 of 16 publishers.
}
\label{tab:article_body_results}
\end{table*}

Table~\ref{tab:multimodal_results} reports metadata, image, and table
extraction performance on the 21-publisher metadata and multimodal
benchmark. Complementing the article-body evaluation, this benchmark
assesses whether PACE can extract structured and multimodal page
elements required by LLM data pipelines. PACE achieves higher performance
than trafilatura on all three targets: 0.755 versus 0.574 for metadata
accuracy, 0.516 versus 0.318 for image F1, and 0.929 versus 0.461 for
table F1. The largest difference is observed in table extraction, where
PACE more than doubles trafilatura's F1.

\begin{table}[t]
\centering
\small
\setlength{\tabcolsep}{4pt}
\begin{tabular}{lccc}
\hline
\textbf{Method} & 
\textbf{Metadata Acc.} $\uparrow$ & 
\textbf{Image F1} $\uparrow$ & 
\textbf{Table F1} $\uparrow$ \\
\hline
PACE & \textbf{0.755} & \textbf{0.516} & \textbf{0.929} \\
trafilatura & 0.574 & 0.318 & 0.461 \\
\hline
\end{tabular}
\caption{
Metadata and multimodal extraction performance across 21 publishers or
domains. Metadata and table scores are computed against GPT-5.4 generated
references; image scores use deterministic in-content image references
parsed from raw HTML. PACE outperforms trafilatura on metadata accuracy,
image F1, and table F1.
}
\label{tab:multimodal_results}
\end{table}

The table extraction results provide the clearest evidence of PACE's
advantage on structured content. As shown in
Table~\ref{tab:table_domain_results}, PACE achieves higher F1 than
trafilatura on all five table-rich domains, with scores ranging from
0.885 to 1.000. These results suggest that template-based extraction with
structured table handling is effective for preserving tabular content,
particularly when pages contain rowspan and colspan structures,
multi-level headers, or row-header stubs. The largest difference is
observed on StockAnalysisFinancials, where PACE achieves 0.919 F1 while
trafilatura scores 0.000, indicating that the generic extractor does not
recover the financial tables in this setting.

\begin{table}[t]
\centering
\small
\setlength{\tabcolsep}{4pt}
\begin{tabular}{lcc}
\hline
\textbf{Domain} & \textbf{PACE} $\uparrow$ & \textbf{trafilatura} $\uparrow$ \\
\hline
W3SchoolsReference & \textbf{1.000} & 0.180 \\
BasketballReferenceStandings & \textbf{0.945} & 0.625 \\
StockAnalysisFinancials & \textbf{0.919} & 0.000 \\
WikipediaLists & \textbf{0.900} & 0.523 \\
BLSEmployment & \textbf{0.885} & 0.443 \\
\hline
\end{tabular}
\caption{
Per-domain table extraction F1 on the five table-rich domains. PACE
consistently achieves higher F1 than trafilatura across all domains,
with particularly large margins on StockAnalysisFinancials and
BLSEmployment.
}
\label{tab:table_domain_results}
\end{table}

For image extraction, PACE achieves an overall image F1 of 0.516,
compared with 0.318 for trafilatura. The advantage is broad but less
uniform than in table extraction: PACE obtains a a higher per-publisher F1
than trafilatura for most image-bearing publishers, ties with trafilatura
on The Telegraph, and obtains a lower F1 on LATimes and The Nation. These
results suggest that publisher-specific decorative-image filtering
improves image extraction overall, while some publishers still require
better distinction among lead images, related thumbnails, author avatars,
and other in-content media.

\subsection{Discussion and Limitations}

Together, the two benchmarks support the central design of PACE. The
article-body benchmark shows that PACE approaches manually engineered
publisher-specific parsers while outperforming scalable non-manual
extractors. The metadata and multimodal benchmark shows that the same
configuration-learning framework can extend to richer extraction targets,
including metadata, images, and tables. In both settings, PACE uses LLMs
during training to learn reusable publisher-specific configurations, then
applies a template-based extraction runtime for scalable inference
without per-page LLM calls.

The metadata and multimodal benchmark relies on GPT-5.4-generated
references rather than human-annotated labels for metadata and table
content. Therefore, the absolute scores should be interpreted as
reference-based estimates rather than measurements against manual ground
truth. However, both PACE and trafilatura are evaluated against the same
references, making the comparison consistent across systems. For image
extraction, we use a deterministic HTML-derived reference because
GPT-5.4 under-extracts image URLs. This reference is transparent and
reproducible, but it remains a heuristic definition of in-content images
rather than a human-labeled annotation.

\section{Conclusion and Future Work}

This paper presented PACE: Publisher-Adaptive Content Extraction via
Agentic Automation, an agentic framework for learning
publisher-specific extraction configurations from representative pages
and user-defined requirements. PACE uses LLMs to interpret extraction
goals, reason over DOM structure, generate page-level specifications,
and aggregate recurring publisher-level patterns. The learned
configurations instantiate a fixed extractor template, enabling scalable
inference without per-page LLM calls.

Across two complementary benchmarks, PACE shows strong article-body
extraction and broader extraction flexibility. On 16 news publishers and
71 held-out pages, PACE achieves the strongest performance among
scalable non-manual methods while approaching manually engineered
publisher-specific parsers. On a 21-publisher metadata and multimodal
benchmark, PACE outperforms trafilatura in extracting metadata, images,
and tables. These results show that agentic configuration learning can
reduce the effort required to support new publishers while preserving
high extraction quality and supporting richer LLM-ready page
representations.

Future work will study how PACE can iteratively improve and scale beyond
independent publisher-specific extractors. Training-page extraction
results can be used to refine analysis prompts, aggregation prompts, and
the extractor template itself. PACE can also be extended beyond news to
product pages, documentation, forums, chat conversations, and enterprise
knowledge bases. Finally, future work should explore shared template
families or more general extractors that reuse common layout patterns
across sites without maintaining an unbounded number of independent
extractors.

\bibliographystyle{unsrt}  
\bibliography{references}

\end{document}